\documentclass{article} 
\usepackage{iclr2025_conference,times}

\usepackage{amsmath,amsfonts,bm}

\def\eqref#1{equation~\ref{#1}}

\def\1{\bm{1}}

\DeclareMathAlphabet{\mathsfit}{\encodingdefault}{\sfdefault}{m}{sl}
\SetMathAlphabet{\mathsfit}{bold}{\encodingdefault}{\sfdefault}{bx}{n}

\newcommand{\bz}{\mathbf{z}}

\usepackage[colorlinks]{hyperref}
\usepackage{url}
\usepackage{graphicx}
\usepackage{xspace}
\usepackage{booktabs}
\usepackage{enumitem}
\usepackage{cleveref}
\usepackage{epsfig}
\usepackage{graphicx}
\usepackage{amsmath}
\usepackage{amssymb}
\usepackage{caption}
\usepackage{multirow}
\usepackage{url}

\usepackage{colortbl} 
\usepackage{xcolor}   
\usepackage{graphicx}
\usepackage{subcaption}
\usepackage{xspace}
\usepackage{multirow}
\usepackage{makecell}
\usepackage{algorithm}
\definecolor{yellow}{rgb}{1, 1, 0.7}
\definecolor{orange}{rgb}{1, 0.85, 0.7}
\definecolor{tablered}{rgb}{1, 0.7, 0.7}
\definecolor{red}{rgb}{1, 0, 0}

\newcommand{\best}{\cellcolor{tablered}}
\newcommand{\sbest}{\cellcolor{orange}}
\newcommand{\tbest}{\cellcolor{yellow}}

\newcommand{\papername}{
Zero-1-to-G\xspace
}
\title{\papername: Taming Pretrained 2D Diffusion Model for Direct 3D Generation}

\author{%
  Xuyi Meng$^{* 1}$, 
  Chen Wang$^{* 1}$, 
  Jiahui Lei$^{1}$, 
  Kostas Daniilidis$^{1}$, 
  Jiatao Gu$^{2}$, 
  Lingjie Liu$^{1}$
  \vspace{0.3em} 
  \\
  ~~~~~~~~~~~~~~~~~~~~~~~~~~~~~~~~~~~~~~~~~~~~~  $^{1}$University of Pennsylvania
  $^{2}$Apple \vspace{0.3em} \\
  ~~~~~~~~~~~~~~~~\url{https://mengxuyigit.github.io/projects/zero-1-to-G/}
}

\iclrfinalcopy 
\begin{document}

\onecolumn{%
\renewcommand\twocolumn[1][]{#1}%
\maketitle

\renewcommand{\thefootnote}{\fnsymbol{footnote}}
\footnotetext[1]{Equal contribution.}

\begin{center}
    \vspace{-20pt}
    \centering
    \captionsetup{type=figure}
    \includegraphics[width=\textwidth]{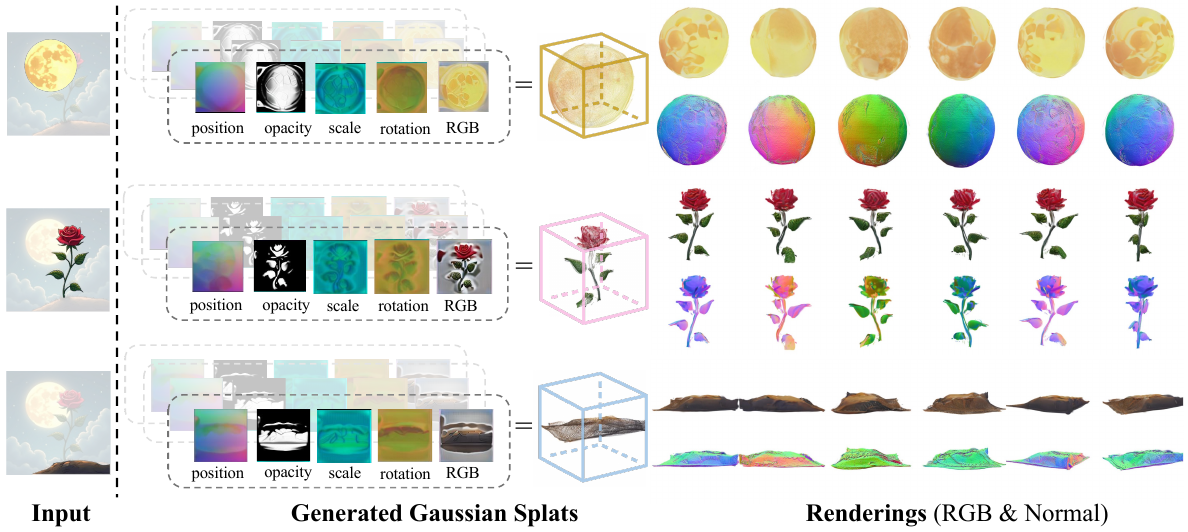}
    \vspace{-15pt}
    \captionof{figure}{\textbf{\papername} tackles direct Gaussian splat generation from single images. By using pretrained 2D diffusion models, we are able to generalize to in-the-wild objects.}
    \label{fig:teaser}
\end{center}%
}

\begin{abstract}
Recent advances in 2D image generation have achieved remarkable quality, largely driven by the capacity of diffusion models and the availability of large-scale datasets. However, direct 3D generation is still constrained by the scarcity and lower fidelity of 3D datasets. In this paper, we introduce \textit{\papername}, a novel approach that addresses this problem by enabling direct single-view generation on Gaussian splats using pretrained 2D diffusion models. 
Our key insight is that Gaussian splats, a 3D representation, can be decomposed into multi-view images encoding different attributes. This reframes the challenging task of direct 3D generation within a 2D diffusion framework, allowing us to leverage the rich priors of pretrained 2D diffusion models. To incorporate 3D awareness, we introduce cross-view and cross-attribute attention layers, which capture complex correlations and enforce 3D consistency across generated splats.
This makes \textit{\papername} the first direct \textbf{image-to-3D} generative model to effectively utilize \textbf{pretrained} 2D diffusion priors, enabling efficient training and improved generalization to unseen objects. Extensive experiments on both synthetic and in-the-wild datasets demonstrate superior performance in 3D object generation, offering a new approach to high-quality 3D generation.

\end{abstract}

\section{Introduction}
\label{sec:intro}

\begin{figure*}[ht!] 
    \centering
    \includegraphics[width=0.98\textwidth]{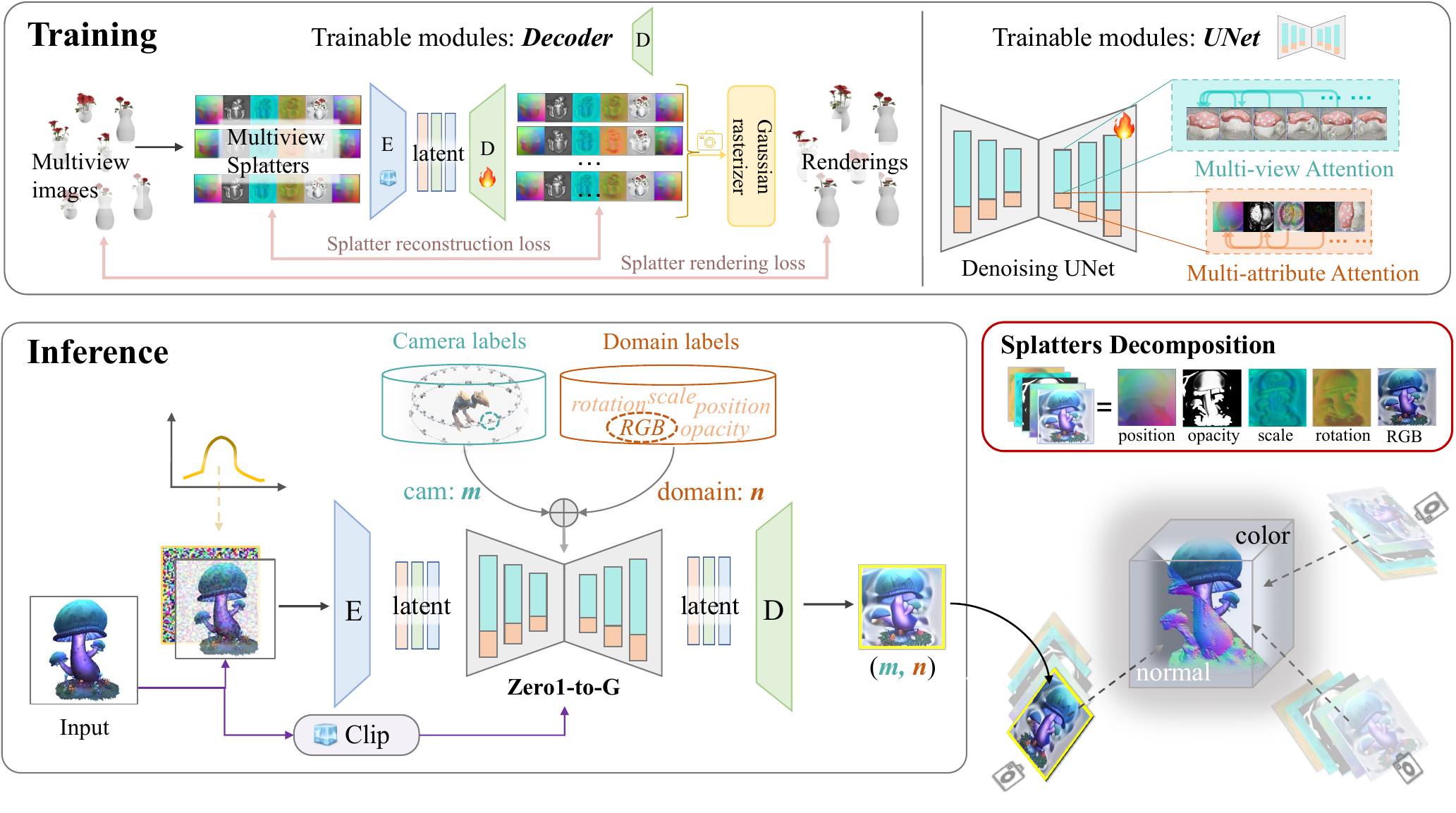}
    \caption{The pipeline of \papername.
    During training, we fine-tune both the VAE decoder (Sec.~\ref{sec:finetune-vae}) and the denoising UNet (Sec.~\ref{sec:mv-and-cd-attention}) of Stable Diffusion.
     At inference time, given a single view input of the target object, each component in the splatter image is generated by conditioning the camera view and attribute switcher.
     The generated set of splatter image components can be directly fused into Gaussian splats (Sec.~\ref{sec:data-decomposition}). Here we show splatter images of 3 views for better illustration, while our main experiments are conducted with 6 views.}
     \vspace{-15pt}
\label{fig:pipeline} 
\end{figure*}

Single image to 3D generation is a pivotal challenge in computer vision and graphics, supporting various downstream applications such as virtual reality and gaming technologies. A primary difficulty lies in managing the uncertainty of unseen regions, as these areas represent a conditional distribution based on the visible portions of a 3D object. Recent advancements in diffusion models~\citep{ho2020denoising, rombach2022high} have demonstrated significant efficacy in capturing complex data distributions within images and videos, prompting researchers to harness these models for single image to 3D generation. Early efforts distilled 3D neural fields from pretrained 2D diffusion models via score distillation~\citep{poole2022dreamfusion, wang2023prolificdreamer}. However, these approaches necessitate per-scene optimization, which is time-consuming and susceptible to multi-faced Janus problems. Subsequent research achieved feed-forward generation by fine-tuning pretrained models to generate multi-view images of the same object~\citep{liu2023zero, shi2023zero123plus, long2023wonder3d, liu2023syncdreamer} and enabling indirect 3D generation through sparse-view reconstruction models~\citep{li2023instant3d, tang2024lgm, xu2024instantmesh}, and Cat3D~\citep{gao2024cat3d} further extends sparse view generation to dense view generation for better reconstruction.
Although these two-stage methods enhance quality and efficiency, they often yield poor geometric fidelity and blurry renderings due to inconsistencies in multi-view images. To circumvent these limitations, recent methodologies have trained diffusion models directly on 3D representations~\citep{liu2023meshdiffusion, chen2023single, zhang2024gaussiancube, he2024gvgen, nichol2022point}, thereby eliminating the reliance on multi-view images. However, direct 3D generation techniques necessitate training from scratch, requiring substantial computational resources and large 3D datasets, which remain scarce—three orders of magnitude less prevalent than 2D data.

In this paper, we propose a novel approach for direct 3D generation that
unites the strengths of both worlds: it leverages the expressive power of 2D diffusion networks while maintaining the 3D structural consistency required for accurate 3D generation.
Our key contribution is bridging the gap between Gaussian splats and natural images typically used in 2D generation tasks.
While the original Gaussian splats consist of 14-channel images encoding various attributes, 
we decompose each of them into multiple 3-channel attribute images
while preserving its 3D information (Sec.~\ref{sec:data-decomposition})
This decomposition also enables efficient latent diffusion training by projecting the splatter images into the latent space of a pretrained VAE, making our method 
directly generate 3D structures within a pretrained 2D diffusion framework. To ensure strong 3D consistency, we introduce cross-view attention layers in Stable Diffusion to enable information exchange between different viewpoints and cross-attribute attention mechanisms to maintain coherence across Gaussian attributes within a splatter image (Sec.~\ref{sec:mv-and-cd-attention}). Additionally, we fine-tune the VAE decoder to address the domain gap between splatter images and natural images, as we observed that splatter image quality is highly sensitive to pixel-level variations (Sec.~\ref{sec:finetune-vae}). 
By leveraging pretrained priors, our method not only generalizes better to unseen objects but also improves training efficiency compared to existing 3D generation methods.

It is important to note that, although we generate multiview splatter images, our method produces more structurally consistent and higher-quality results compared to traditional two-stage multiview 3D generation approaches. In two-stage methods~\citep{xu2024instantmesh, tang2024lgm}, strong pixel-level consistency is required in the first stage to ensure accurate reconstruction in the second stage, which is often difficult to achieve. Moreover, the first stage operates independently of the second, lacking coordination between the two, thereby exacerbating inconsistencies. In contrast, our approach directly generates the final 3D representation in a single stage, eliminating the need for pixel-level correspondence between multiview splatter images. Since splatter images can contain redundant information, and their spatial positions do not necessarily map to the final 3D positions, this single-stage process offers greater flexibility and robustness, ensuring a more consistent final 3D structure.


Overall, our contributions can be summarized as below:
\begin{itemize}[leftmargin=1cm, rightmargin=1cm]
    \item We present \textit{Zero-1-to-G}, a novel direct 3D generative model for Gaussian splats that achieves excellent 3D consistency and superior rendering quality. 
    \item We observe that Gaussian splats, as a 3D representation, can be decomposed into a set of 2D images of different views and attributes, making them inherently compatible with 2D image generation frameworks.
    \item Through decomposition and transformation of splatter images, we use 2D diffusion models for direct 3D generation with proper fine-tuning, unleash the power of pretrained 2D diffusion for training efficiently and better generalization towards in-the-wild data.

    
\end{itemize}

\section{Related Works}
\label{sec:related_works}

\noindent\textbf{Optimization-based 3D Generation}
Dreamfusion~\citep{poole2022dreamfusion} and subsequent works~\citep{lin2023magic3d, chen2023fantasia3d, wang2023prolificdreamer} utilize a pretrained text-to-image diffusion model to optimize a 3D representation through score distillation. DreamGaussian~\citep{tang2023dreamgaussian} significantly reduces training time by optimizing Gaussian splats. However, score distillation-based methods still require minutes of optimization per scene, as they must compare renderings with diffusion outputs from various viewpoints, which limits their generation speed. Additionally, these methods lack a clear understanding of geometry and viewpoint, resulting in multi-face problems.

\noindent\textbf{Direct 3D Generation}
Significant efforts have been made to directly train diffusion models on various 3D representations, including point clouds~\citep{luo2021diffusion, zhou20213d, nichol2022point, jun2023shap}, meshes~\citep{liu2023meshdiffusion}, and neural fields~\citep{chen2023single, shue20233d, muller2023diffrf}. However, these methods are typically constrained to category-level datasets and often struggle to generate high-quality assets. More recent approaches have begun encoding 3D assets into more compact latent representations~\citep{zhang20233dshape2vecset, lan2024ln3diff, zhao2023michelangelo, zhang2024clay, hong20243dtopia, dong2024gpld3d}, enabling diffusion models to be trained more efficiently and enhancing generalization capabilities. Despite these advancements, direct 3D generative models are still primarily trained on synthetic 3D datasets like Objaverse~\citep{deitke2024objaverse}, which may hinder their ability to effectively handle more in-the-wild inputs.

More closely related to our work are GVGen~\citep{he2024gvgen} and GaussianCube~\citep{zhang2024gaussiancube}, which investigate training diffusion models on Gaussian splats. Both approaches acknowledge the challenges of directly learning diffusion models from Gaussian splats and propose organizing the Gaussian points into a more structured volume. Different from their methods, we adopt a multi-view splatter image representation, enabling us to train diffusion models on Gaussian splats directly with high efficiency.

\noindent\textbf{Finetuning Pretrained Diffusion models} 
To enhance the 3D awareness of 2D pretrained diffusion models, MVDream~\citep{shi2023mvdream} aintegrates cross-view attention layers and fine-tunes them to produce multi-view renderings. 
As an application, LGM~\citep{tang2024lgm} and InstantMesh~\citep{xu2024instantmesh} utilize a pre-trained single-view to multi-view 2D diffusion model, transforming the single view generation problem to a multi-view reconstruction task. These methods adopt various compact 3D representations to adapt to the sparse view 2D input. This has greatly pushed the 3D generation to more complex data. However, this approach is constrained by the performance of the multi-view diffusion model, which often lacks strict multi-view consistency. This inconsistency can result in poor geometry and blurry textures in the final 3D reconstructions.

Some other works further leverage the power of pretrained diffusion models to generate data beyond the domain of natural images.
For instance, JointNet~\citep{zhang2023jointnet} uses two diffusion models for joint RGB and depth prediction. Marigold~\citep{ke2023repurposing} fine-tunes pretrained diffusion models for monocular depth estimation from single image input.
Wonder3D~\citep{long2023wonder3d} generates both multi-view RGB and normal maps from a unified diffusion model equipped with a domain switcher. Similarly, GeoWizard~\citep{fu2024geowizard} predicts depth and normals from a single image using cross-domain geometric self-attention to maintain geometric consistency.
Following this line of work to adapting diffusion prior to images of other domains, 
our method aims to generate Gaussian splats represented as splatter images using pretrained 2D diffusion models, thereby enhancing the ability of direct 3D generation to tackle in-the-wild images.

\noindent\textbf{3D Generation with Reconstruction-based methods} 
Researchers have opted to train reconstruction-based models for highly efficient 3D generation~\citep{hong2023lrm,tochilkin2024triposr,woo2024harmonyview,zou2023triplane,xu2023dmv3d}.
LRM~\citep{hong2023lrm} and TripoSR~\citep{tochilkin2024triposr} introduced a transformer-based model that directly output a triplane from a single image. The model is trained on million-scale data by comparing the renderings of the triplane with ground truth using regression-based loss. TriplaneGaussian~\citep{zou2023triplane} further used a hybrid triplane-Gaussian representation to greatly accelerate the rendering of the generated 3D assets.
However, the main drawback of regression-based methods is their failure to account for the uncertainty in single-view to 3D generation. 
Although GECO~\citep{wang2024geco} attempts to address this issue by distilling knowledge from multi-view diffusion models into a feedforward model, its results remain limited by the quality of the generated multi-view images.

\noindent\textbf{Direct 3D Generation with 2D Diffusion}
Concurrent works \citep{yan2024object} and \citep{elizarov2024geometry} also propose methods of generating 3D with 2D diffusion models
but use UV atlases to encode geometry and texture, whereas we employ Splatter Images~\citep{szymanowicz23splatter} as the 3D representation.
Specifically, Omage~\citep{yan2024object} uses a 12-channel UV atlas, requiring training from scratch without leveraging pretrained 2D diffusion priors, limiting its generalization ability. 
On the other hand, GIMDiffusion~\citep{elizarov2024geometry} decomposes the UV atlas into separate geometry maps and albedo textures to match the 3-channel output, but the use of pretraiend 2D diffusion model are limited to albedo generation. 
Still, both approaches focus on text-to-3D generation while we focus more on single-view image-to-3D reconstruction,
and their reliance on mesh representations limits flexibility to model real-world data containing complex backgrounds, while our approach offers greater adaptability to diverse scenarios.

\section{Methods}
\label{sec:methods}

Our method \papername is a single stage direct 3D generation: given single view input $\mathbf{I}$, \papername generates the corresponding 3D representation $\mathbf{z}$, where $z = \{z_i|i=1,...,N\}$ multiple Splatter Images under $\mathbf{N}$ camera views.

To harness the power of large-scale pretrained 2D diffusion models for direct 3D generation, we represent each 3D object as a set of multi-view Splatter Images~\citep{szymanowicz23splatter}. In Sec.~\ref{sec:data-decomposition}, we detail our decomposition process, converting each multi-view splatter image into five 2D attribute images corresponding to RGB color, scale, rotation, opacity, and position. This decomposition allows us to effectively leverage the priors of 2D pretrained diffusion models to learn the underlying 3D object distribution (Sec.~\ref{sec:mv-and-cd-attention}). Furthermore, we fine-tune the VAE decoder to enhance the rendering quality of the decoded Splatter Images (Sec.~\ref{sec:finetune-vae}).


\subsection{Representing 3D objects as 2D images}
\label{sec:data-decomposition}

Gaussian splats can be rearranged into regular grids of $ {H} \times {W} $ called Splatter Image~\citep{szymanowicz23splatter}, with 14 channels stacking 5 Gaussian splat attributes.
By applying the transformation detailed in~\Cref{sec:appendix/implemetation-details}, 
each attribute can be represented as an RGB image that explicitly models the object's appearance or geometry. 
Example visualization of such transformation are shown in \Cref{fig:pipeline}.
Our key observation is that, each attribute image is well modeled within the distribution of pretrained 2D diffusion model, suggesting that these priors could be effectively utilized for generating Splatter Images, which serve as a 3D representation.
Inspired by prior works leveraging pretrained diffusion priors for 3D-rich domain-specific images such as normals or depth~\citep{ke2023repurposing, long2023wonder3d, fu2024geowizard}, we extend this approach to generate five attribute images of Gaussian splats.

Generating ground-truth splatters for training is a critical step in our pipeline. A straightforward approach involves fitting splatters for each object based on multi-view renderings. However, the per-object fitting-based method introduces excessive high-frequency signals in the Splatter Images. Such high-frequency artifacts lead to significant appearance changes after encoding and decoding with the pretrained VAE, indicating that the pretrained 2D diffusion model struggles to capture these signals effectively.
Thus, we opted to use the splatters produced by neural networks because inherently they exhibit smoother characteristics. 
Specifically, we fine-tune the \textbf{reconstruction module} of LGM~\citep{tang2024lgm} that outputs splatters from the ground truth 2D renderings of the training set.
This fine-tuning generates Splatter Images that can be effectively encoded and decoded by the pretrained VAE, allowing us to leverage the pretrained priors efficiently.
It is important to emphasize that our method is not inherently bounded by LGM because
we train and infer the network with the same set of data, this ensures the model fits its parameters on the training set and does not have potential generalizability issues. 

\begin{figure}[h!]
\centering
\includegraphics[width=0.99\textwidth]{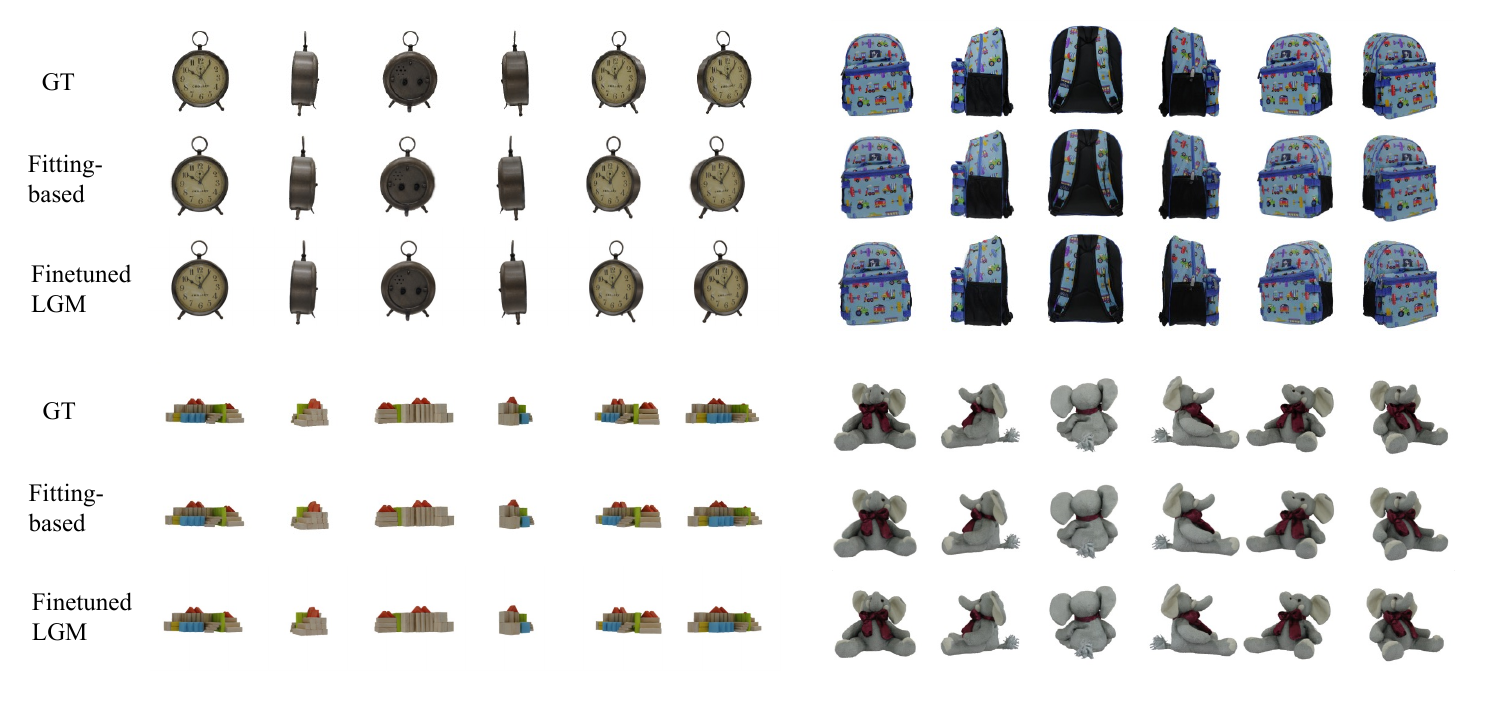}
\caption{Qualitative comparison between rendering results from splatters acquire through fitting-based methods for each object, versus the splatters predicted by our fine-tuned LGM in a feed-forward manner.}
\vspace{-10pt}
\label{fig:lgm-finetune}
\end{figure}

Also, using network to generate ground truth splatters is much more efficient than per-scene fitting without suffering the rendering quality, 
a comparison between our fine-tuned LGM and the fitting-based method on the test dataset can be found in \Cref{fig:lgm-finetune}, validating our choice of using LGM reconstruction module in generating high-quality ground-truth Splatter Images.

\begin{figure}[h!]
\centering
\includegraphics[width=0.99\textwidth]{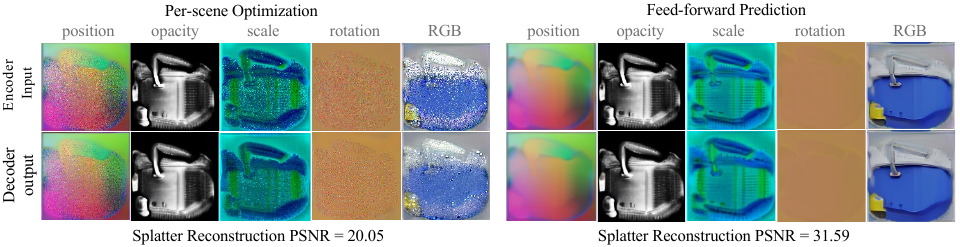}
\caption{VAE encoding and decoding comparison with per-scene optimized splatters and feed-forward predicted splatters.}
\vspace{-10pt}
\label{fig:encode-decode}
\end{figure}




\subsection{Direct 3D Generation via 2D Pretrained Diffusion}
\label{sec:mv-and-cd-attention}
With the decomposition discussed in Sec~\ref{sec:data-decomposition}, we are ready to directly learn a 3D generative model on Gaussian splats represented by a set of multi-view attribute images.
The distribution of our objects represented as Splatter Images, denoted as $p(\bz)$, is modeled as a joint distribution of 6 fixed camera views and 5 splatter attributes. Given a set of fixed camera viewpoints $\left\{\boldsymbol{\pi}_1, \boldsymbol{\pi}_2, \cdots, \boldsymbol{\pi}_K\right\}$ and condition input image $y$:
\begin{align}
\label{eq:conditional-diffusion}
    p(\mathbf{z}) = p(\bz^{(1:K, 1:N)} | y) = p_{\text{pos, op, sc, rot, rgb}} \left( \{\bz_{\text{pos}}^{1:K}, \bz_{\text{op}}^{1:K}, \bz_{\text{sc}}^{1:K}, \bz_{\text{rot}}^{1:K}, \bz_{\text{rgb}}^{1:K} \} \mid y \right)
\end{align}
where ${\{\text{pos, op, sc, rot, rgb}\}}$ are the 5 attributes of the splatter image.

Namely, we can naturally utilize the pretrained 2D diffusion models to generate these attribute images since they are already formatted as 3-channel images. However, original Stable Diffusion can only generate single-view images, while our goal is to learn the joint distribution across multiple views and attributes.
To ensure that the generated 5 multi-view attribute images are coherent and represent the same object, 
we insert additional attention layers into the pretrained diffusion UNet to jointly model both the cross-view and cross-attribute distribution of our decomposed Splatter Images.

\noindent\textbf{Modeling Multi-view Distribution}  
Prior works have approached multi-view diffusion either by reshaping the batch dimension into a token dimension and applying self-attention~\citep{shi2023mvdream, liu2023syncdreamer, long2023wonder3d, liu2024part123}, or by spatially concatenating multi-view images to form a larger image, which directly maps the latent distribution to a multi-view distribution~\citep{shi2023zero123plus}. We choose the former approach for its flexibility in reshaping data for both cross-view and cross-attribute attention mechanisms. This design allows for efficient information exchange among different views, where tokens corresponding to the same attribute from different views are concatenated for self-attention. This facilitates our model's ability to learn a consistent multi-view distribution for each Gaussian attribute.

\noindent\textbf{Modeling Multi-Attribute Distribution}
Building on the work of \citep{long2023wonder3d}, we utilize an attribute switcher to specify which attribute the network should generate. To maintain consistency across generated images that represent different attributes of the same object, we employ an attention mechanism to capture the interactions between images taken from the same viewpoint but corresponding to different attributes. 

Specifically, we introduce additional self-attention modules to model the cross-attribute correlations, where tokens representing all attributes from the same viewpoint are combined and processed using standard scaled dot-product attention.

\noindent\textbf{Training Loss} 
During training, we organize each view and attribute of a splatter image within the batch dimension and apply independently sampled Gaussian noise. In each attention block, we alternately apply multi-view attention and multi-attribute attention to enhance the model's ability to learn complex correlations.

The forward process of our diffusion model is directly extended from the original DDPM~\citep{ho2020denoising}, which is
\begin{equation}
\label{eq:mv_forward}
q(\mathbf{z}^{(1:K, 1:N)}_{1:T}|\bz_0^{(1:K, 1:N)}) = \prod_{t=1}^T q(\mathbf{z}^{(1:K, 1:N)}_t|\mathbf{z}^{(1:K,1:N)}_{t-1}) 
= \prod_{t=1}^T \prod_{k=1}^K \prod_{n=1}^N q(\bz^{(k, n)}_t|\bz^{(k, n)}_{t-1}),
\end{equation}

And the reverse process will be
\begin{align}
p_\theta(\bz_{0:T}^{(1:K, 1:N)}) &= p(\bz^{(1:K, 1:N)}_T) \prod_{t=1}^{T} p_\theta(\bz^{(1:K, 1:N)}_{t-1}|\bz^{(1:K, 1:N)}_{t}) \\
&= p(\bz^{(1:K, 1:N)}_T) \prod_{t=1}^{T} \prod_{k=1}^{K} \prod_{n=1}^{N} p_\theta(\bz^{(k, n)}_{t-1}|\bz^{(1:K, 1:N)}_{t}),
\label{eq:mv_reverse1}
\end{align}
where $p_\theta(\bz^{(k, n)}_{t-1}|\bz^{(1:K, 1:N)}_{t}) = \mathcal{N}(\bz_{t-1}^{(k, n)}; \mathbf{\mu}_\theta^{(k, n)}(\bz_t^{(1:K, 1:N)}, t), \sigma_t^2 \mathbf{I}).$. The definition of the Gaussian mean for the reverse process is defined as:
\begin{equation}
    \mathbf{\mu}_\theta^{(k, n)}(\bz^{(1:K, 1:N)}_t, t) = \frac{1}{\sqrt{\alpha_t}} \left( \bz_t^{(k,n)} - \frac{\beta_t}{\sqrt{1 - \bar{\alpha}_t}} \mathbf{\epsilon}_\theta^{(k,n)}(\bz^{(1:K, 1:N)}_t, t) \right),
\end{equation}


The corresponding loss function for multi-view and multi-domain modeling is as follows:
\begin{equation}
    \ell = \mathbb{E}_{t, \mathbf{x}^{(1:K, 1:N)}_0, k, n, \mathbf{\epsilon}^{(1:K, 1:N)}} \left[ \|\mathbf{\epsilon}^{(k, n)} - \mathbf{\epsilon}_\theta^{(k, n)}(\bz^{(1:K, 1:N)}_t, t)\|_2^2 \right],
\end{equation}
where \( \mathbf{\epsilon}^{(k, n)} \) is the Gaussian noise added to attribute \( n \) for the \( k \)-th view, and \( \mathbf{\epsilon}_\theta^{(k, n)} \) is the model's noise prediction for attribute $n$ in the $k$-th view.

\subsection{VAE Decoder Fine-tuning}
\label{sec:finetune-vae}

The pretrained VAE of Stable Diffusion is initially trained on natural images. While directly utilizing this VAE can reconstruct visually appealing Splatter Images, it does not guarantee high-quality RGB renderings and they often exhibit noticeable artifacts. These artifacts arise from two main factors: (1) each pixel in the splatter image corresponds to a Gaussian splat, meaning that even minor changes in pixel values can significantly affect the final rendering, and (2) Splatter Images contain high-frequency details that are challenging for the VAE to recover accurately.

To address these challenges while leveraging pretrained knowledge, we freeze the VAE encoder and fine-tune the decoder using splatter rendering losses. This approach refines the VAE for our task while preserving its pretrained priors, similar to practices in image and video diffusion models. For example, Stable Diffusion fine-tunes its VAEs on human image datasets to improve facial reconstruction, and video diffusion models~\citep{blattmann2023align} fine-tune VAE decoders to enhance temporal consistency. Likewise, our fine-tuning enhances the VAE's understanding of 3D information in Splatter Images, ensuring high-fidelity reconstruction and rendering.

Specifically, the rendering loss comprises two components: MSE loss and LPIPS loss, defined as follows:

\begin{equation}
\begin{aligned}
\mathcal{L}_{\text{rgb}} &= \mathcal{L}_{\text{MSE}} + \mathcal{L}_{\text{LPIPS}}
\end{aligned}
\end{equation}

The overall objective function of decoder finetuning is defined as:
\begin{equation}
\begin{aligned}
\mathcal{L}_{\text{decoder}} &= \mathcal{L}_{\text{splatter}} + \mathcal{L}_{\text{normal}} + \mathcal{L}_{\text{rgb}} + \mathcal{L}_{\text{mask}}
\end{aligned}
\end{equation}

where $\mathcal{L}_{\text{splatter}}$ denotes the reconstruction loss of splatter image itself, while
$\mathcal{L}_{\text{normal}}$, $\mathcal{L}_{\text{rgb}}$ and $\mathcal{L}_{\text{mask}}$ are all about the 
renderings of the reconstructed splatter, denoting 
cosine similarity loss of rendered normals, the sum of all losses of rendered images, and the binary cross-entropy loss of the rendered masks.

\begin{figure*}[t]
\centering
\includegraphics[width=\textwidth]{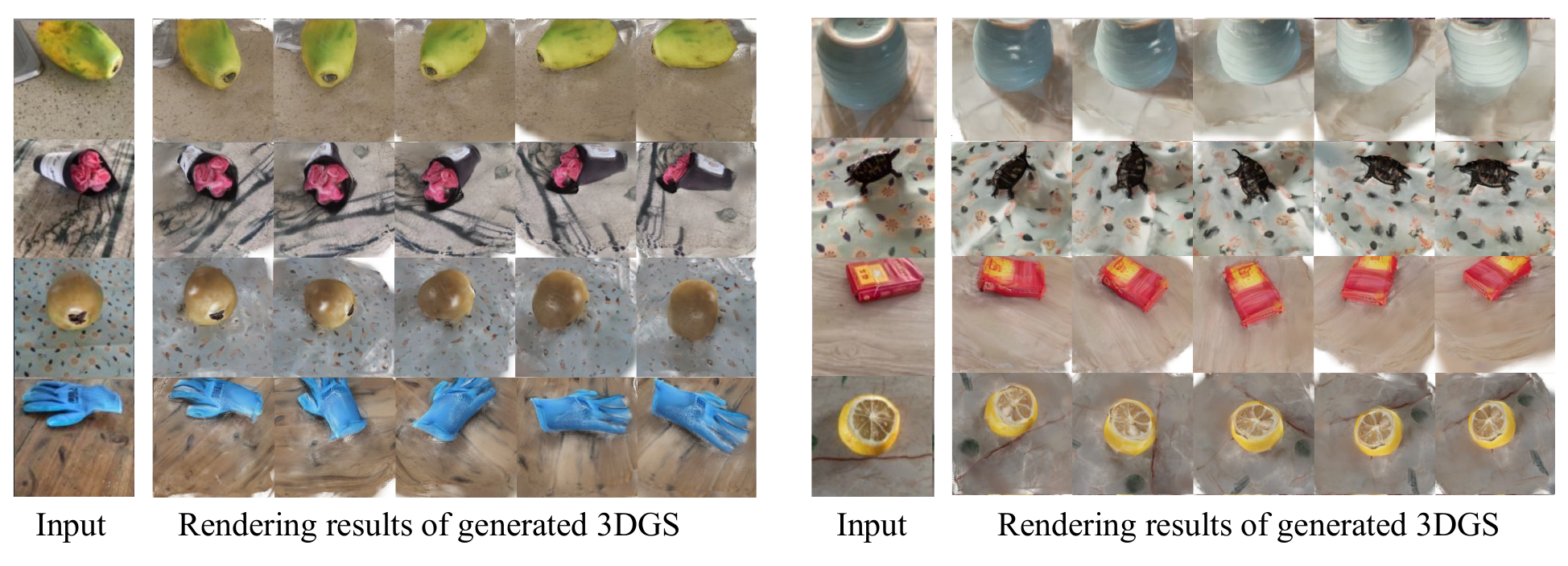}
\vspace{-15pt}
\caption{RGB and normal renderings of more examples on MVImgNet dataset.}
\label{fig:mvimagenet}
\end{figure*}

\begin{figure}[h!]
\centering
\includegraphics[width=0.99\textwidth]{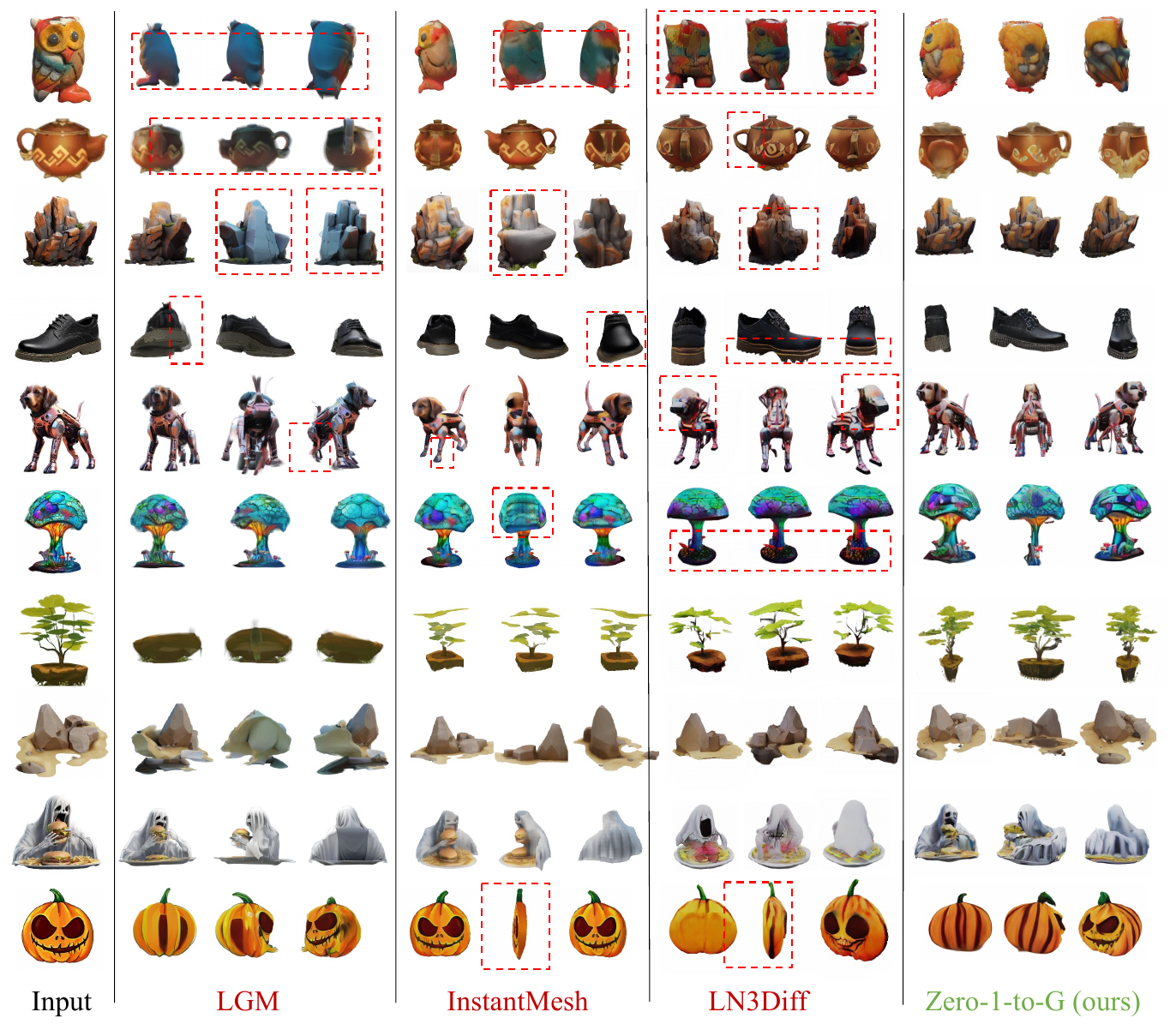}
\vspace{-5pt}
\caption{Qualitative comparison with, LGM, InstantMesh, LN3Diff on in-the-wild data.}
\label{fig:exp-baseline-in-the-wild}
\vspace{-5pt}
\end{figure}

\section{Experiments}
\label{sec:experiments}

\subsection{Implementation Details}
\noindent\textbf{Dataset}
We train on the G-buffer Objaverse~\citep{qiu2024richdreamer} dataset, which consists of approximately 262,000 objects sourced from Objaverse~\citep{deitke2024objaverse}. Each object in the dataset is rendered from 38 viewpoints, with additional normal and depth renderings provided. For generating the ground truth splatter images, we use the first viewpoint as the input condition, along with five additional views at the same elevation and azimuth angles of {30°, 90°, 180°, 270°, and 330°} to comprehensively cover the full 360 degrees. We only use the RGB and normal renderings for the supervision of decoder finetuning and Gaussian Splats prediction model.

\noindent\textbf{Model Training and Inference} 
We initialize our model from Stable Diffusion Image Variations. Following Wonder3D~\citep{long2023wonder3d}, our training includes two stages. In the first stage, we only train multi-view attention,
and in the second stage, we add one more cross-domain attention layer for training, and together fine-tune the multi-view attention layer learned in the first stage. 
For the first stage, we use a batch size of 64 on 4 NVIDIA L40 GPUs for 13k iterations, which takes about 1 day. For the second stage, we use a batch size of 64 on 8 NVIDIA L40 GPUs for 30k iterations, which takes about 2 days.
For decoder fine-tuning, we use a total batch size of 64 on 8 NVIDIA L40 GPUs for 20k iterations. The second stage of training takes about 2 days. During inference, we use $\text{cfg}=3.5$ and our method can generate Gaussian splats per object in $8.7$ seconds on a single NVIDIA L40 GPU.





\subsection{Evaluation Protocol}
\noindent\textbf{Dataset and Metrics} 
Following prior works~\citep{liu2023syncdreamer, liu2023one2345++, wang2024geco}, we conduct quantitative comparisons using the Google Scanned Objects (GSO) dataset~\citep{downs2022google}. Specifically, we utilize a randomly selected subset of 30 objects from the GSO dataset, including a variety of everyday items and animals, as in SyncDreamer~\citep{liu2023syncdreamer}. For each object, a conditioning image is rendered at a resolution of 
$512 \times 512$ with an elevation angle of $10^\circ$.  Evaluation images are then generated at evenly spaced 
$30^\circ$ azimuthal intervals around the object, keeping the elevation constant.

To assess the quality of novel view synthesis, we report standard metrics such as PSNR, SSIM~\citep{wang2004image}, and LPIPS~\citep{zhang2018unreasonable}. 
Additionally, we evaluate the geometry of our generated outputs using Chamfer Distance (CD). Please refer to~\Cref{tab:comparison} for details.

Beyond the GSO dataset, we also evaluate our approach on in-the-wild images to demonstrate its robustness and generalizability (\Cref{fig:exp-baseline-in-the-wild}).

\noindent\textbf{Baselines} We compare our methods against several recent approaches across different categories. For reconstruction-based methods, we include TriplaneGaussian~\citep{zou2023triplane} and TripoSR~\citep{tochilkin2024triposr}. In the realm of direct 3D generation, we compare with LN3Diff~\citep{lan2024ln3diff}. Finally, for two-stage methods transitioning from single-image to multi-view to 3D, we include InstantMesh~\citep{xu2024instantmesh} and LGM~\citep{tang2024lgm}.


\begin{table}[h!]
\caption{Quantitative comparison between our methods and other baselines on the GSO dataset.}
\vspace{-15pt}
\begin{center}
\begin{tabular}{ccccc}
\toprule
Methods & PSNR $\uparrow$ & SSIM $\uparrow$ & LPIPS $\downarrow$ & CD $\downarrow$ \\ \cmidrule(lr){1-1}\cmidrule(lr){2-5}
TriplaneGaussian~\citep{tang2023dreamgaussian}  & $17.80$ & $0.811$ & $0.216$  & $0.0440$    \\ 
TripoSR~\citep{tochilkin2024triposr}  & $17.32$ & $0.804$ & $0.217$  & $0.0423$   \\ 
LGM~\citep{tang2024lgm} & $17.01$ &  $0.793$ & $0.199$ & $0.0621$    \\ 
InstantMesh~\citep{xu2024instantmesh}    & \tbest $18.15$  & \tbest $0.810$   & \sbest $0.179$ & \tbest $0.0419$  \\ 
LN3Diff~\citep{lan2024ln3diff} & $16.30$ &  $0.786$ & $0.241$ & $0.0637$ \\ \cmidrule(lr){1-5}
Ours-fast (10 steps) & \sbest $19.03$ &  \sbest $0.812$ & \tbest $0.182$ & \sbest $0.0396$\\
Ours (35 steps)     & \best $\mathbf{19.40}$      & \best $\mathbf{0.818}$   & \best$\mathbf{0.178}$ & \best $\mathbf{0.0390}$  \\ \bottomrule
\end{tabular}
\vspace{-15pt}
\end{center}
\label{tab:comparison}
\end{table}

\subsection{Results}


\noindent\textbf{Qualitative Comparison} \Cref{fig:exp-baseline-in-the-wild} showcases rendering results of \papername against baselines on in-the-wild inputs. Two-stage methods like LGM~\citep{tang2024lgm} and InstantMesh~\citep{xu2024instantmesh} often face quality and consistency issues due to reliance on multi-view generation. LGM frequently produces smooth, artifact-prone textures (e.g., blue backgrounds in the first and third examples) and inconsistent 3D Gaussians with “floaters” (fourth and fifth examples). InstantMesh offers more consistent renderings but introduces smoothness and grid-like texture artifacts
from its reconstruction process. 
Both approaches are limited by upstream multi-view generation, resulting in flawed geometry and inconsistency (last example).
In contrast, our method employs inherent 3D representations, achieving accurate geometry and consistent renderings.
Compared to other direct 3D generation method LN3Diff~\citep{lan2024ln3diff}, it struggles with fine-grained textures and produces oversmoothed geometry. Training from scratch limits its quality, generalization, and performance on unseen objects due to constrained resources and datasets.
In contrast, by leveraging pretrained 2D diffusion priors and use one-stage direct generation, our approach delivers high-fidelity 3D renderings with consistent geometry and texture, effectively handling in-the-wild input images (\Cref{fig:exp-baseline-in-the-wild}), as well as real-world data with backgrounds (\Cref{fig:mvimagenet}).



\noindent\textbf{Quantitative Comparison} The quantitative results on the GSO dataset, presented in \Cref{tab:comparison}, show that \papername consistently outperforms all baselines across all metrics. Reconstruction-based methods, like TriplaneGaussian and TripoSR, struggle with sharp predictions for unseen regions due to their deterministic nature. Two-stage methods, such as InstantMesh, perform reasonably well but are still limited by sparse multi-view images. Direct 3D methods like LN3Diff underperform due to the lack of pretrained priors.

\noindent\textbf{Training Efficiency}
By leveraging pretrained diffusion priors, our method reduces training time and resource requirements. We complete training in just 3 days using only 8 NVIDIA L40 GPUs, which is more efficient compared to previous two-stage and direct 3D generation methods, as detailed in \Cref{tab:efficiency}. This efficiency highlights the advantage of integrating 2D priors for direct 3D generation, reducing the need for extensive computational resources.


\begin{figure}[h!]
\centering
\includegraphics[width=0.99\textwidth]{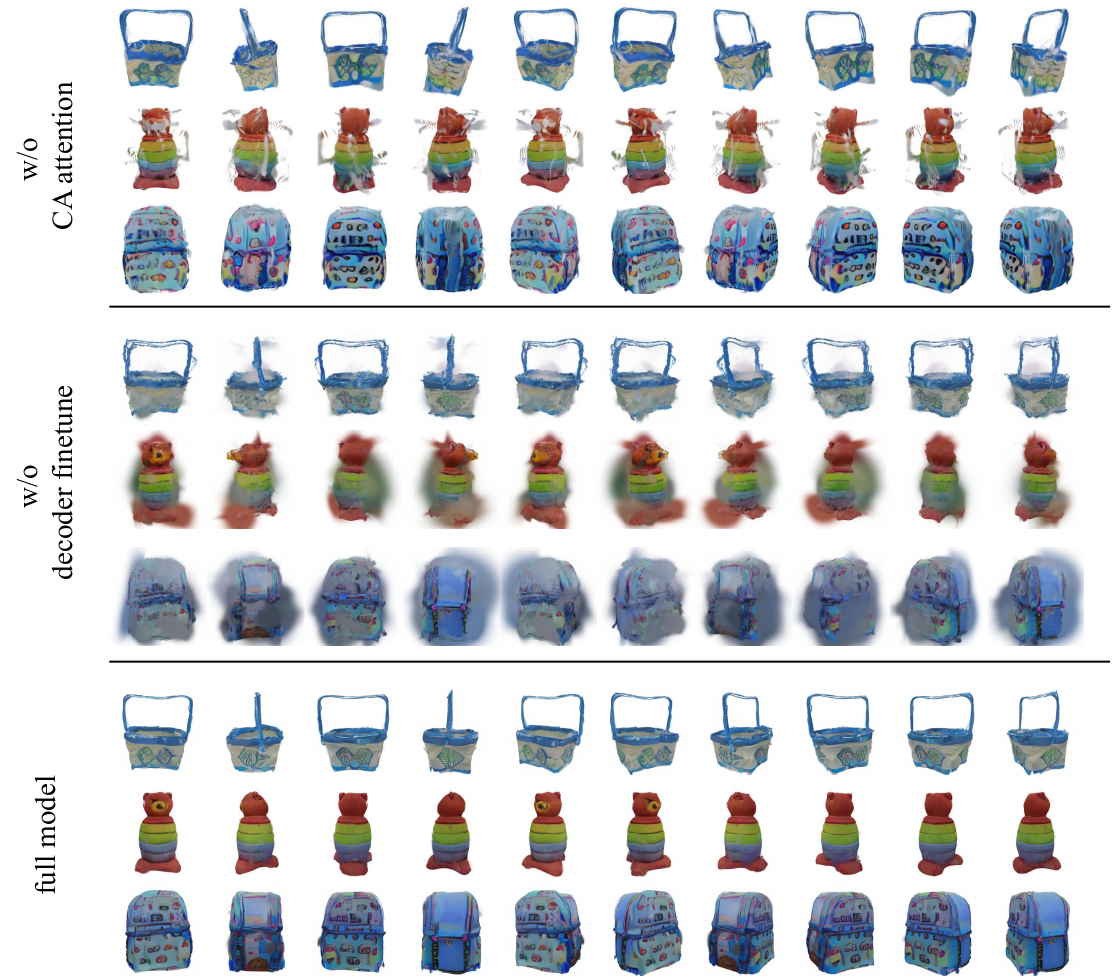}
\caption{Ablation study on GSO dataset.}
\label{fig:ablation}
\end{figure}

\subsection{Ablation Study}
\label{subsec:ablation}
\noindent\textbf{VAE Decoder Finetuning} Without fine-tuning VAE decoder, although the decoded splatter image visually looks fine, the renderings exhibit noticeable artifacts. Since each pixel represents a Gaussian splat and the decoder cannot capture high-frequency areas, well-reconstructed splatter images don't necessarily ensure good renderings.

\noindent\textbf{Cross-attribute Attention} We can see from \Cref{fig:ablation}, if we don't use cross-attribute attention, the renderings of the Gaussian splats have many floaters and the textures become blurry, this is because that different attributes of the same Gaussian splat are not well aligned.

\noindent\textbf{No Diffusion Prior} The use of diffusion prior is essential in our model. To verify this, we conducted training with the same StableDiffusion UNet architecture but using random initialization, and we also added the same cross-view attention layers to the UNet as in our method. We can see that without using the prior, the training of the model cannot converge to meaningful results using the same data and training iterations.

\begin{table}[h!]
\vspace{-10pt}
\caption{Comparison of training efficiency with other baseline methods. 
For LGM and InstantMesh, we only count the reconstruction part. More time and resources are needed to train their multi-view generation model.}
\vspace{-15pt}
\begin{center}
\begin{tabular}{ccc}
\toprule
Methods & Training Time $\downarrow$ & GPUs $\downarrow$ \\ \cmidrule(lr){1-1}\cmidrule(lr){2-3}
LGM (reconstruction module)   & 4 days & 32 * A100 (80G)   \\ 
InstantMesh (reconstruction module)  & 12 days   & 16 * H800 (80G)  \\ 
LN3Diff    & 7 days & 8 * A100 (80G) \\ 
Ours      &  3 days & 8 * L40 (48G)  \\ \bottomrule
\end{tabular}
\end{center}
\vspace{-5pt}
\label{tab:efficiency}
\end{table}

\begin{table}[h!]
\caption{Abalation study on module design, inference with GSO dataset.}
\vspace{-10pt}
\begin{center}
\begin{tabular}{cccc}
\toprule
Model Components & PSNR $\uparrow$ & SSIM $\uparrow$ & LPIPS $\downarrow$ \\ \cmidrule(lr){1-1}\cmidrule(lr){2-4}
w/o diffusion prior & $9.39$ & $0.592$ & $0.722$ \\
w/o decoder fine-tuning  & $17.13$ & $0.775$ & $0.272$  \\ 
w/o cross-attribute attention & $17.26$ & $0.767$ & $0.237$ \\
Full Model & $\mathbf{19.40}$ &$\mathbf{0.818}$ & $\mathbf{0.178}$  \\

\bottomrule
\end{tabular}
\end{center}
\vspace{-10pt}
\label{tab:ablation}
\end{table}


\section{Conclusion}

In this work, we introduce a novel framework that leverages 2D diffusion priors for direct 3D generation by decomposing Gaussian splats into multi-view attribute images. This decomposition preserves the full 3D structure while efficiently mapping it to 2D images, enabling fine-tuning of pretrained Stable Diffusion models with cross-view and cross-attribute attention layers. Our approach significantly reduces computational costs compared to other direct 3D generation methods. 
By bypassing the stringent requirement for multi-view image consistency in two-stage approaches, we generate more accurate 3D geometry and produce higher-quality renderings through a single-stage diffusion process. Furthermore, our method exhibits stronger generalization capabilities than existing direct 3D generation techniques due to the use of diffusion priors, offering a more efficient and scalable solution for 3D content creation.

\paragraph{Limitations \& Future Works}
Despite achieving superior reconstruction metrics and strong generalization to in-the-wild data, our method has some limitations. First, our inference speed is not as fast as regression models, as each splatter must be generated through diffusion. A potential improvement would be to integrate a diffusion distillation~\citep{song2023consistency, wang2024geco, gu2023boot} to reduce denoising steps. Second, we do not currently disentangle material and lighting conditions, leading to highlights and reflections being baked into the Gaussian splat texture. Future work could address this by incorporating inverse rendering to better predict non-Lambertian surfaces.

\bibliography{iclr2025_conference}
\bibliographystyle{iclr2025_conference}

\appendix
\section{Appendix}

\subsection{Implementation Details}
\label{sec:appendix/implemetation-details}
\noindent\textbf{Splatter Image Transformation} \\
Each attribute, except opacity, possesses three degree-of-freedoms, which align gracefully with the 3 channels of the RGB space.
The following illustrates the detailed operations to convert each attribute into an RGB image
\begin{itemize}[leftmargin=1cm, rightmargin=1cm]
    \item RGB: the RGB attribute naturally lies in the RGB space, no conversion is needed.
    \item Position: We normalize the 3D object in the bounding box $[-1, 1]$ and use the 3D coordinates of each Gaussian as position attribute.
    \item Scale: The raw scale value spans from $1e{-15}$ to $1e{-2}$, so directly converting the 3D scale to RGB space using the min-max value for the whole dataset will result in most regions being zeros due to the significant difference in power. The value distribution also does not match the normal distribution, making it difficult for diffusion models to learn effectively. We thus convert the raw scale values to log-space and clamp the minimum values to $-10$, as we found Gaussian splats with scales smaller $1e{-10}$ will have negligible effects on the final rendering.
    \item Rotation: We first convert the 4-dimensional quaternion to 3-dimensional axis angle, then normalize it to $[-1, 1]$.
    \item Opacity: We directly duplicate the single channel to 3 channels, and average the predicted 3-channel image to get the final opacity prediction.
\end{itemize}

\noindent\textbf{UNet Fine-tuning} 
When fine-tuning the StableDiffusion UNet, for both stages, we use a constant learning rate of $1e{-4}$ with a warmup of the first 100 steps. We use the Adam optimizer for both stages and the betas are set to $(0.9, 0.999)$. For classifier-free guidance, we drop the condition image with a probability of $0.1$.

\noindent\textbf{LGM Fine-tuning} To obtain the splatter image ground truth for our training, as mentioned in Sec.~\ref{sec:data-decomposition}, we fine-tune LGM~\citep{tang2024lgm} to take as input 6 multi-view renderings of the G-Objaverse dataset and output splatter images of 2D Gaussian splatting~\citep{huang20242d}. The training objective is to compare the splatter renderings with ground truth images using MSE and LPIPS loss. We also use cosine similarity loss between ground truth normals and rendered normals. We fine-tune LGM for $30$k iterations with a batch size of $32$ on 8 NVIDIA L40 GPUS, which takes about 1 day.


\subsection{More Results}

\begin{figure}[htbp]
\centering
\includegraphics[width=\textwidth]{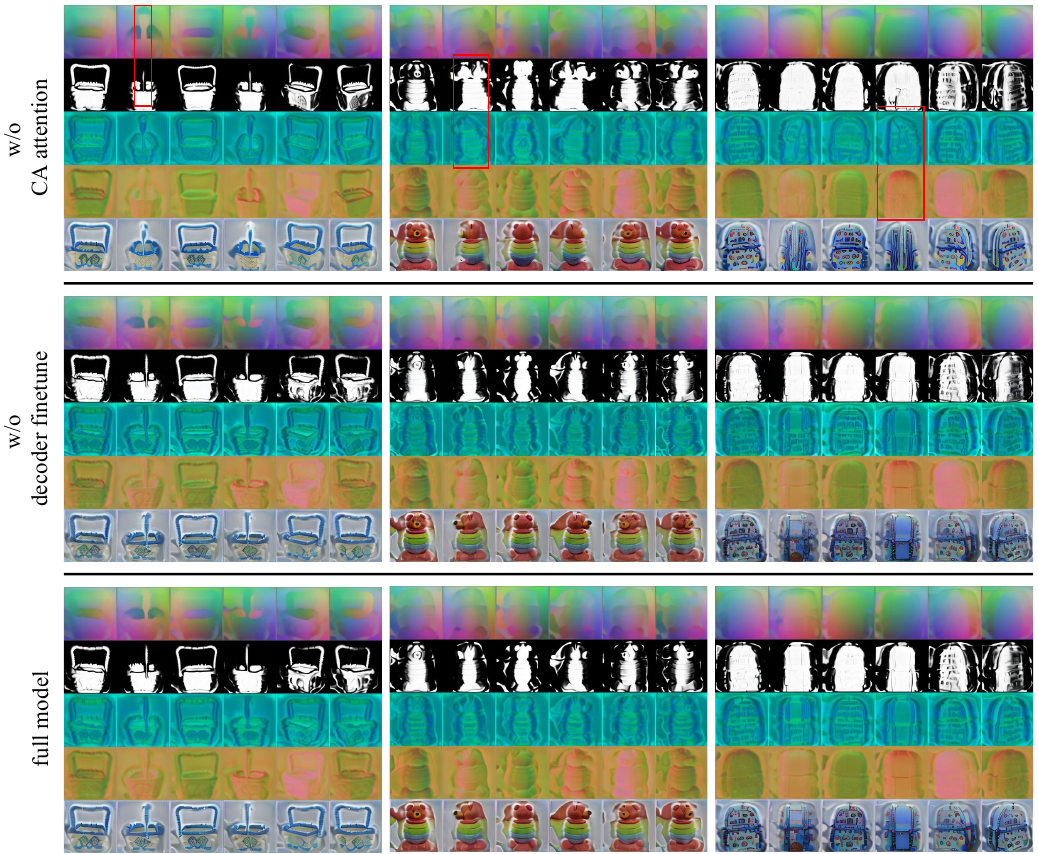}
\caption{Splatter visualization of ablation study.}
\label{fig:ablation-splatter}
\end{figure}

\noindent\textbf{More visualizations on ablation} \Cref{fig:ablation-splatter} shows the splatter image visualization of ablation study. We can see that without cross-attribute attention, there are obvious misalignments of different domains of the splatter images. Without decoder fine-tuning, although the splatter image is visually good, the rendering is not satisfying because Gaussian splats are sensitive to the value changes in the pixels. Fine-tuning the decoder can greatly improve the rendering quality.

\begin{figure}[h!]
\centering
\includegraphics[width=1\textwidth]{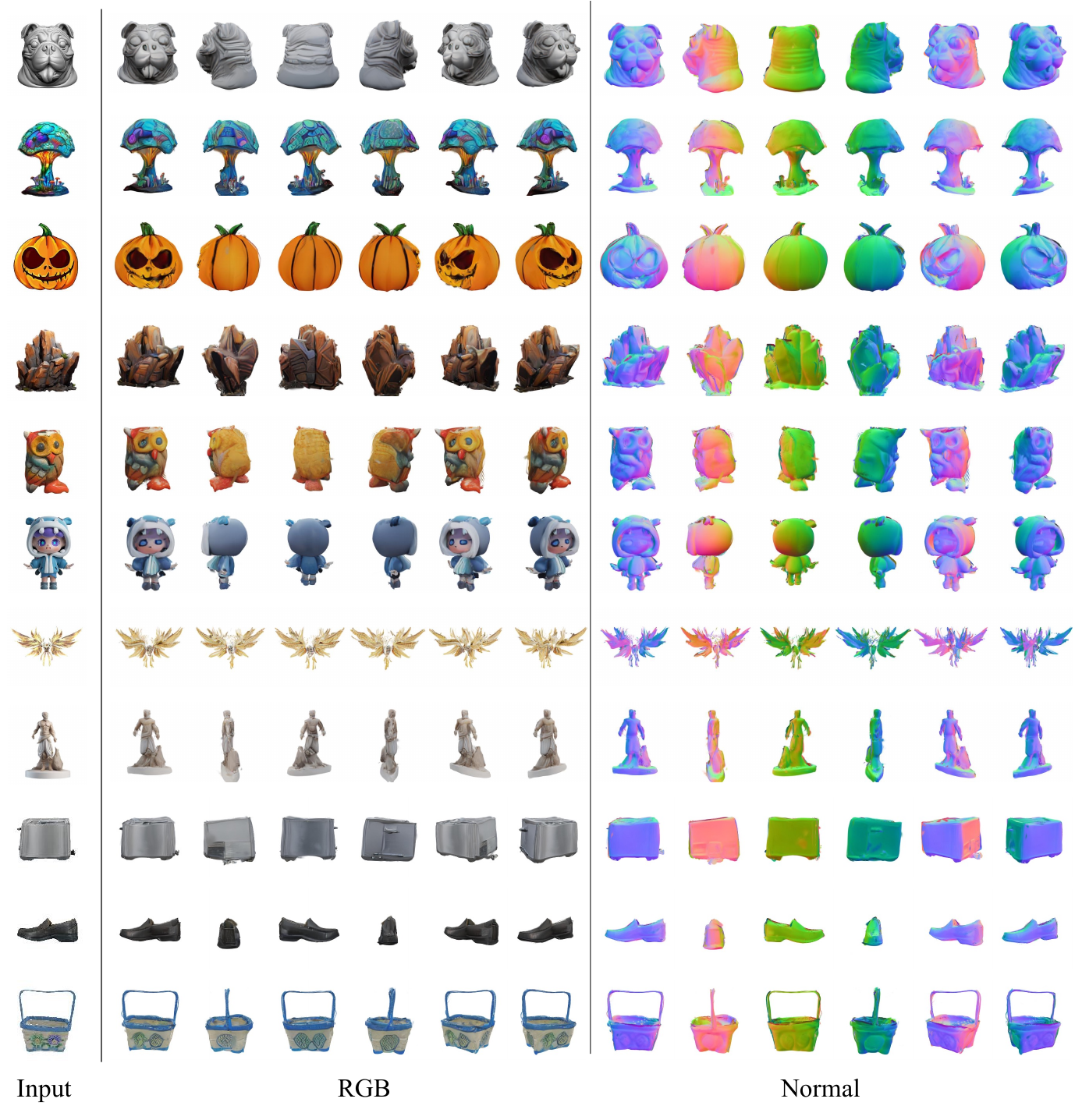}
\caption{RGB and normal renderings of more examples on in-the-wild and GSO datasets.}
\label{fig:more-results}
\end{figure}

\noindent\textbf{More examples} More RGB and normal renderings can be found in \Cref{fig:more-results} and \Cref{fig:mvimagenet}.

\begin{figure}[h!]
\centering
\includegraphics[width=\textwidth]{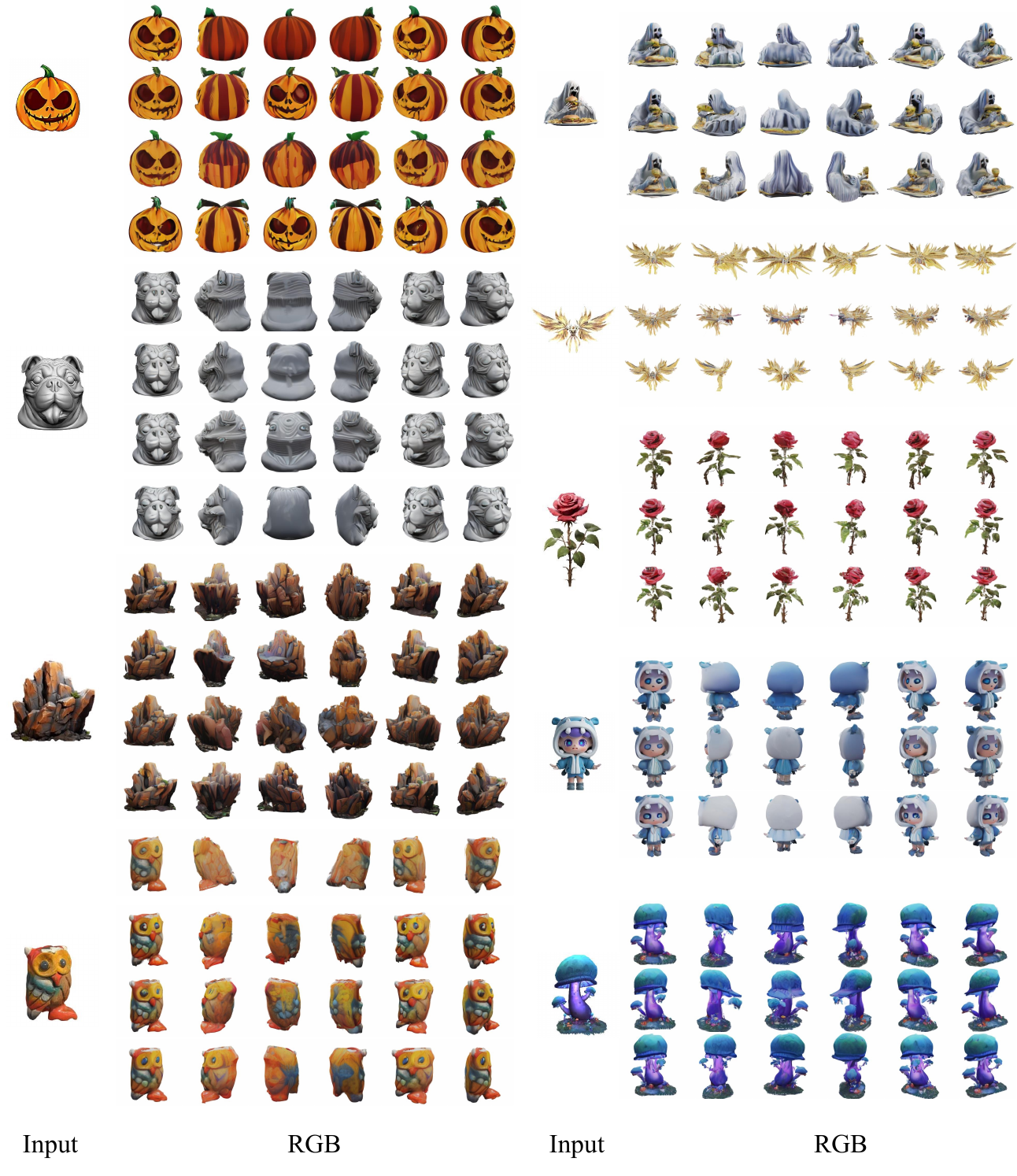}
\caption{Generative 3D model with various geometry and texture given the same condition image, which shows the strong generative ability of our model.}
\vspace{-15pt}
\label{fig:diversity}
\end{figure}

\noindent\textbf{Diversity} Since we model the unseen viewpoints with diffusion models, our results can generate diverse results given the same input~\Cref{fig:diversity}.

\end{document}